\documentclass[twoside,leqno,twocolumn]{article}  
\usepackage{ltexpprt} 
\usepackage{graphicx}

\begin{document}
\title{Neural Discourse Modeling of Conversations}
\author{John M. Pierre, Mark Butler, Jacob Portnoff, Luis Aguilar
\\
Voise Inc
\\
 \texttt{jpierre,mbutler,jportnoff,laguilar@voiseinc.com}}
\date{}
\maketitle
%
\begin{abstract}

Deep neural networks have shown recent promise in many language-related tasks
such as the modeling of conversations.
We extend RNN-based sequence to sequence models to capture
the long range discourse across many turns of conversation. 
We perform a sensitivity analysis on how much additional context affects performance, 
and provide quantitative and qualitative evidence that these models 
are able to capture discourse relationships across multiple utterances.
Our results quantifies how adding an additional RNN layer for modeling
discourse improves the quality of output utterances and providing more of the previous
conversation as input also improves performance.
By searching the generated outputs for specific discourse markers 
we show how neural discourse models can exhibit increased coherence
and cohesion in conversations.

\end{abstract}
\section{Introduction}
\label{intro}

Deep neural networks (DNNs) have been successful in modeling many aspects of 
natural language including word meanings \cite{mikolov}\cite{le},
machine translation \cite{sutskever},
syntactic parsing \cite{vinyals_grammar}, language modeling \cite{jozefowicz} and
image captioning \cite{vinyals_caption}.
Given sufficient training data, DNNs are
highly accurate and can be trained end-to-end without the need for 
intermediate knowledge representations or explicit feature extraction.
With recent interest in conversational user interfaces such as virtual
assistants and chatbots, the application of DNNs to facilitate
meaningful conversations is an area where more progress is needed.
While sequence to sequence models based on recurrent
neural networks (RNNs) have shown initial promise in creating intelligible
conversations \cite{vinyals_conversation}, it has been noted that more work is needed
for these models to fully capture larger aspects of human communication including
conversational goals, personas, consistency, context, and word knowledge.
 
Since discourse analysis considers language at
the conversation-level, including its social and psychological context, 
it is a useful framework for guiding the extension of end-to-end neural conversational models.
Drawing on concepts from discourse analysis such as \textit{coherence}
and \textit{cohesion} \cite{halliday}, 
we can codify what makes conversations more intelligent 
in order to design more powerful neural models that reach beyond the
sentence and utterance level.
For example, by looking for features that indicate deixis, anaphora,
and logical consequence in the machine-generated utterances we can benchmark the level of coherence and
cohesion with the rest of the conversation, and then make improvements to
models accordingly.

In the long run, if neural models can encode the long-range structure
of conversations, they may be able to express conversational discourse similar to the way the human brain does, without the need for
explicitly building formal representations of discourse theory into the model.

To that end, we explore RNN-based sequence to
sequence architectures that can 
capture long-range relationships between multiple utterances in conversations
and look at their ability to exhibit discourse relationships.
Specifically, we look at 1) a baseline RNN encoder-decoder with attention mechanism and 2) a model with an
additional discourse RNN that encodes a sequence of multiple utterances.

Our contributions are as follows:
\begin{itemize}
\item We examine two RNN models with attention mechanisms to model
  discourse relationships across different utterances
that differ somewhat compared to what has been done before
\item We carefully construct controlled experiments 
to study the relative merits of different models on multi-turn conversations
\item We perform a sensitivity analysis on how the amount of context
  provided by previous utterances affects model performance
\item We quantify how neural conversational models display coherence by measuring the prevalence of specific
syntactical features indicative of deixis, anaphora, and logical consequence. 
\end{itemize}

\section{Related Work}
\label{relatedwork}

Building on work done in machine translation, sequence to sequence models based on RNN encoder-decoders were
initially applied to generate conversational outputs given a single previous
message utterance as input\cite{shang}\cite{vinyals_conversation}.
In \cite{sordoni} several models were presented that included a
``context'' vector (for example representing another previous utterance)
that was combined with the message utterance via various encoding
strategies to initialize or bias a single decoder RNN.
Some models have also included an additional RNN tier to capture the
context of conversations. For example, 
\cite{serban} includes a hierarchical ``context RNN'' layer to summarize
the state of a dialog, while
\cite{yao} includes an RNN ``intension network'' to model conversation
intension for dialogs involving two participants speaking in turn. 
Modeling the ``persona'' of the participants in a conversation by
embedding each speaker into a $K$-dimensional embedding was shown
to increase the consistency of conversations in \cite{li_persona}.

Formal representations such as Rhetorical Structure Theory (RST) \cite{mann}
have been developed to identify discourse structures in written text. 
Discourse parsing of cue phrases \cite{marcu} and coherence modeling
based on co-reference resolution of named-entities \cite{barzilay}\cite{kibble} have been applied to tasks such as 
summarization and text generation. Lexical chains \cite{morris} and
narrative event chains \cite{chambers} provide directed graph models of text
coherence by looking at thesaurus relationships and subject-verb-temporal
relationships, respectively. 
Recurrent convolutional neural networks have been used to classify
utterances into discourse speech-act labels \cite{kalchbrenner} and
hierarchical LSTM models have been evaluated for generating coherent
paragraphs in text documents \cite{li} .

Our aim is to develop end-to-end neural conversational models that exhibit
awareness of discourse without needing a formal representation of
discourse relationships.

\subsection{Models}
\label{models}

Since conversations are sequences of utterances and utterances are sequences of words, it is natural to use models based on an RNN encoder-decoder
to predict the next utterance in the conversation given $N$
previous utterances as source input.
We compare two types of models: \textbf{seq2seq+A}, which applies an attention mechanism
directly to the encoder hidden states, and \textbf{Nseq2seq+A}, which
adds an additional RNN tier with its own attention mechanism to model discourse relationships between $N$ input utterances.

In both cases the RNN decoder predicts the output utterance and
the RNN encoder reads the sequence of words in each input utterance. 
The encoder and decoder each have their own vocabulary
embeddings.

\begin{figure*}[t]
\centerline{\includegraphics[width=12cm]{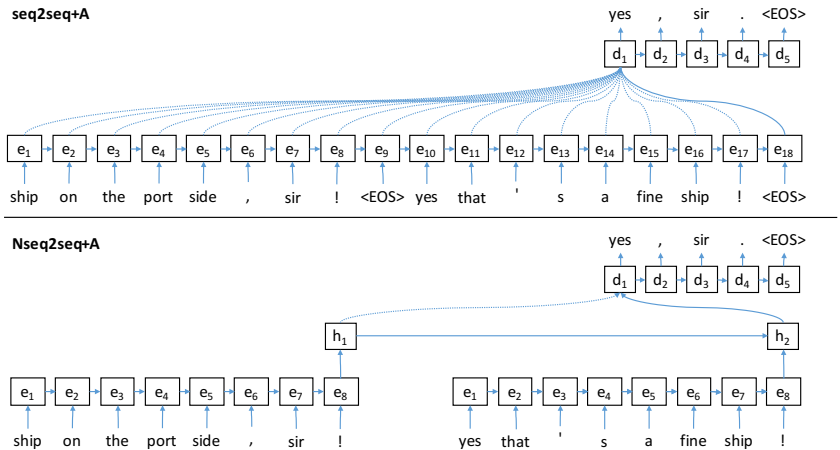}}
\caption{Schematic of \textbf{seq2seq+A} and \textbf{Nseq2seq+A}
  models for multiple turns of conversation. An attention mechanism is
  applied either directly to the encoder RNN or to an intermediate
  discourse RNN.}
  \label{schematic}
\end{figure*}

As in \cite{vinyals_grammar} we compute the attention vector at each
decoder output time step $t$
given an input sequence $(1,...,T_{A})$ using:
\begin{eqnarray}
u_{i}^{t} & = & v^{T}tanh(W_{1} h_{i} + W_{2} d^{t}) \nonumber \\
a_{i}^{t} & = & softmax(u_{i}^{t}) \nonumber \\
c^{t} & = & \sum_{i=1}^{T_{A}} a_{i}^{t} h_{i} \nonumber
\end{eqnarray}
where the vector $v$ and matrices $W_{1}$, and $W_{2}$ are learned parameters.
$d^{t}$ is the decoder state at time $t$ and is concatenated with $c^{t}$ to
make predictions and inform the next time step. In \textbf{seq2seq+A} the $h_{i}$ are the
hidden states of the encoder $e_{i}$, and for \textbf{Nseq2seq+A} they are the
$N$ hidden states of the discourse RNN (see Fig. \ref{schematic}.)
Therefore, in
\textbf{seq2seq+A} the attention mechanism is applied at the
word-level, while in \textbf{Nseq2seq+A} attention is applied at the utterance-level.

\subsection{seq2seq+A}
\label{seq2seq}

As a baseline starting point we use an attention mechanism to help model
the discourse by a straightforward adaptation of 
the RNN encoder-decoder conversational model discussed in \cite{vinyals_conversation}.
We join multiple source utterances using the \textit{EOS} symbol as a delimiter, and feed them into the
encoder RNN as a single input sequence. 
As in  \cite{sutskever}, we reversed the order of the tokens
in each of the individual utterances but preserved the order of
the conversation turns. 
The attention mechanism is able to make connections to any of the words used in earlier 
utterances as the decoder generates each word in the output response.

\subsection{Nseq2seq+A}
\label{Nseq2seq}

Since conversational threads are ordered sequences of utterances,
it makes sense to extend an RNN encoder-decoder by adding another
RNN tier to model the discourse as the turns of the conversation progress.
Given $N$ input utterances, the RNN encoder is applied to each 
utterance one at a time as shown in Fig. \ref{schematic} (with tokens
fed in reverse order.)
The output of the encoder from each of the input utterances forms
$N$ time step inputs for the discourse RNN.
The attention mechanism is then applied to the $N$ hidden states of the
discourse RNN and fed into the decoder RNN.
We also considered a model where the output of the encoder is
also combined with the output of the discourse RNN and fed into the
attention decoder,
but found the purely hierarchical architecture performed better.

\subsection{Learning}
\label{learning}

For each model we chose identical optimizers, hyperparameters, etc. in our experiments
 in order to isolate the impact of specific differences in the network
architecture, also taking computation times and available GPU resources into account. 
It would be straightforward to perform
a grid search to tune hyperparameters, try LSTM cells, increase layers
per RNN, etc. to further improve
performance individually for each model beyond what we report here.

For each RNN we use one layer of Gated Recurrent Units (GRUs) with
512 hidden cells.
Separate embeddings for the encoder and decoder, each with dimension
512 and vocabulary size of 40,000, are trained on-the-fly without using predefined
word vectors.

We use a stochastic gradient descent (SGD) optimizer with $L2$ norms
clipped at $5.0$, an initial learning rate of $0.5$, and a learning rate decay
factor of $0.99$ is applied when needed. We trained with mini-batches of 64
randomly selected examples, and ran training for approximately 10
epochs until validation set loss converged.

\section{Experiments}
\label{experiments}
We first present results comparing 
our neural discourse models trained on a large set of conversation
threads based on the OpenSubtitles dataset \cite{tiedemann}.
We then examine how our models are able to produce
outputs that indicate enhanced coherence by searching for discourse markers.

\subsection{OpenSubtitles dataset}
\label{opensubtitles}

A large-scale dataset is important if we want to model all the
variations and nuances of human language.
From the OpenSubtitles corpus we created a training set and validation
set with 3,642,856 and 911,128 conversation fragments,
respectively\footnote{The training and validation
sets consisted of $320M$ and $80M$ tokens, respectively}.
Each conversation fragment consists of 10 utterances from
the previous lines of the movie dialog leading up to a target utterance.
The main limitation of the OpenSubtitles dataset is that it is derived
from closed caption style subtitles, which can be noisy, do not include labels for
which actors are speaking in turn, and do not show conversation
boundaries from different scenes.

We considered cleaner datasets such as the Ubuntu dialog corpus
\cite{lowe}, Movie-DiC dialog corpus \cite{banchs}, and SubTle corpus
\cite{ameixa} but found they all contained orders of magnitude fewer
conversations and/or many fewer turns per conversation on average.
Therefore, we found the size of the OpenSubtitles dataset outweighed the benefits of
cleaner smaller datasets. This echoes a trend in neural networks
where large noisy datasets tend to perform better than small clean datasets.
The lack of a large-scale clean dataset of conversations is an open
problem in the field.

\subsection{Results}
\label{results}

We compared models and performed a sensitivity analysis by varying the
number of previous conversation turns fed into the encoder during
training and evaluation. 

\begin{table*}[!htbp]
\caption{Results on OpenSubtitles dataset. Perplexity vs. number of
  previous conversation turns.}
\label{perplexity}
\begin{center}
\begin{tabular}{cll}
\hline
\textbf{Previous conversation turns} & \textbf{seq2seq+A} & \textbf{Nseq2seq+A}  \\
\hline
$N=1$ & 13.84$\pm$0.02 & 13.71$\pm$0.03 \\
$N=2$ & 13.49$\pm$0.03 & 13.40$\pm$0.04 \\
$N=3$ & 13.44$\pm$0.05 & 13.31$\pm$0.03 \\
$N=5$ & -              & 13.14$\pm$0.03 \\
$N=7$ & -              & 13.08$\pm$0.03 \\
\hline
\end{tabular}
\end{center}
\end{table*}

\begin{figure*}[t]
\centerline{\includegraphics[width=11cm]{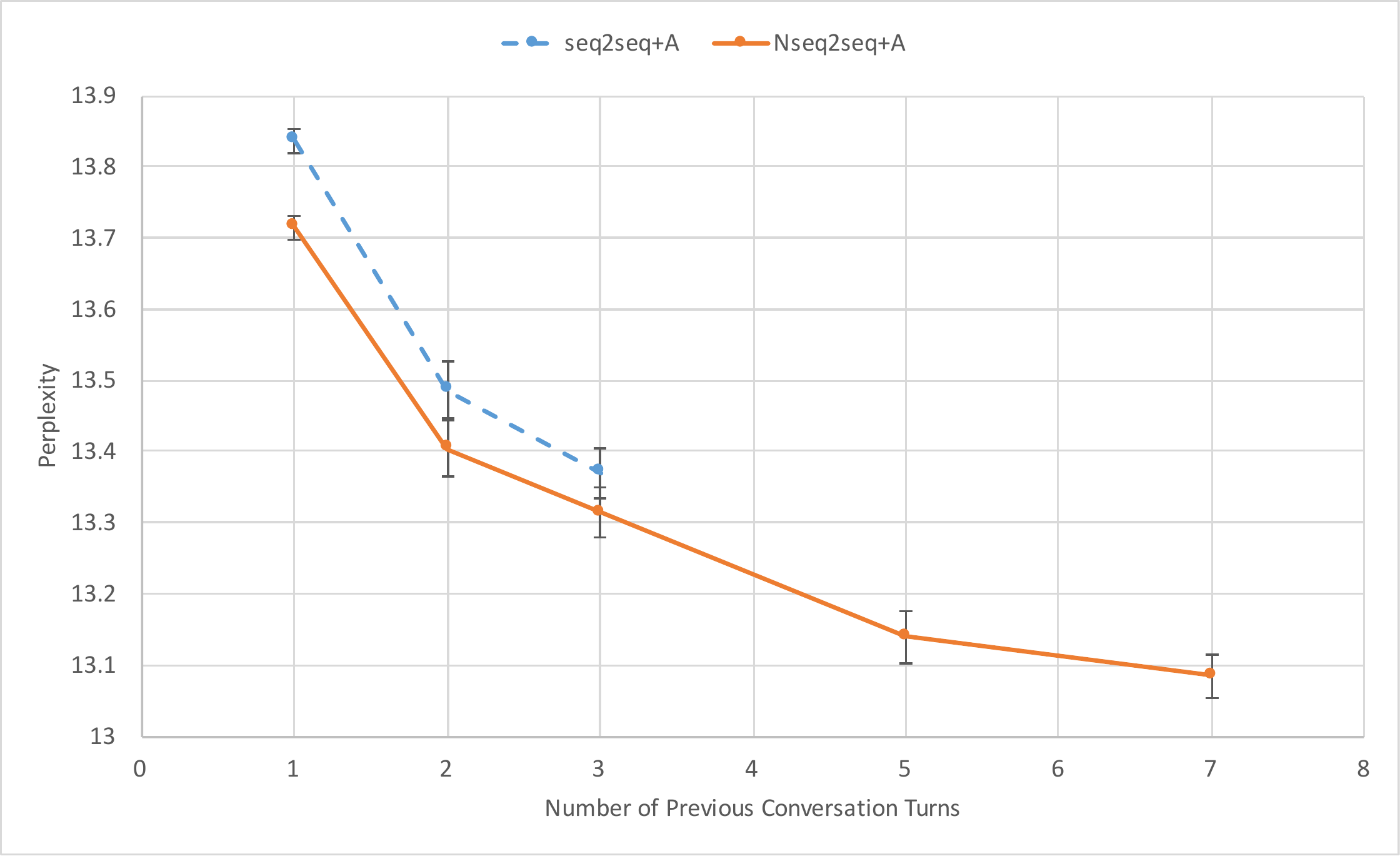}}
\caption{Sensitivity analysis of perplexity vs. number of previous conversations turns.}
\label{sensitivity}
\end{figure*}

In Table \ref{perplexity} we
report the average perplexity\footnote{We use perplexity as our performance metric, because it is simple to
compute and correlates with human judgements, though it has well-known limitations.} on the validation set at convergence for each model.
For $N=1,2,3$ we found that \textbf{Nseq2seq+A}
shows a modest but significant performance improvement over the baseline
\textbf{seq2seq+A}. We only ran \textbf{Nseq2seq+A} on larger values
of $N$, assuming it would continue to outperform.

In Fig. \ref{sensitivity} we show that increasing the amount
of context from previous conversation turns significantly improves
model performance, though there appear to be diminishing returns.

\subsection{Discourse analysis}
\label{discourse}

Since a large enough dataset tagged with crisp discourse
relationships is not currently available, we seek a 
way to quantitatively compare relative levels of coherence and cohesion.
As an alternative to a human-rated evaluation we performed
simple text analysis to search for specific discourse markers \cite{fraser}
that indicate enhanced coherence in the decoder output as follows:
\begin{itemize}
\item \textbf{Deixis:} contains words or phrases\footnote{here, there, then, now, later, this, that} referring
  to previous context of place or time
\item \textbf{Anaphora:} contains pronouns\footnote{she,
    her, hers, he, him, his, they, them, their, theirs}
referring to entities mentioned in previous utterances
\item \textbf{Logical consequence:} starts with a cue
  phrase\footnote{so, after all, in addition, furthermore, therefore,
    thus, also, but, however, otherwise, although, if, then} 
forming logical relations to previous utterances
\end{itemize}

\begin{table*}[!htbp]
\caption{Discourse analysis of \textbf{Nseq2seq+A} decoder
  output. Likelihood of discourse markers vs. number of previous
conversation turns used as input.}
\label{cohesion}
\begin{center}
\begin{tabular}{llllll}
\hline
& $N=1$ & $N=2$ & $N=3$ & $N=5$ & $N=7$ \\
\hline
deixis & 4.0\%  & 3.4\% & 16.3\% & 5.1\% & 5.0\% \\
anaphora & 4.4\%  & 6.1\% & 9.9\% & 7.2\% & 9.3\% \\
logical consequence & 0.03\% & 0.05\% & 0.08\% & 0.34\% & 0.12\% \\
\hline
\end{tabular}
\end{center}
\end{table*}

In Table \ref{cohesion} we show how $N$, the number of 
previous conversation turns used as input,
 affects the likelihood that these discourse
markers appear in the decoder output. The percentage of output
utterances containing discourse markers related to deixis, anaphora, and logical
consequence are reported from a sample of 100,000 validation set
examples.

In general, we find that more context leads to a higher likelihood of
discourse markers indicating that long-range discourse relationships are indeed
being modeled. 
The results show a potentially interesting sensitivity to the value of $N$, require further study, and are
likely to be dependent on different conversational styles and domains.

\subsection{Examples}
\label{examples}

In Table \ref{cohesion} we show a few examples comparing decoder outputs
of the \textbf{Nseq2seq+A} model using either 1 or 5 previous
conversation turns as input. Qualitatively we can see that this
neural discourse model is capable of producing increased cohesion when provided with
more context.

\begin{table*}[!htbp]
\caption{Example responses comparing \textbf{Nseq2seq+A} with either 1
  or 5 previous conversation turns as input. Output words exhibiting cohesion with
  previous conversation turns are indicated in bold.}
\label{responses}
\begin{center}
\begin{tabular}{|l|l|}
\hline
\textbf{Previous Lines of Conversation Input} & \textbf{Decoder Output} \\
\hline
\hline
 can ' t see a number plate , even . & \textit{i ' m sorry .} \\
\hline
then there ' s this . & \\
five streets away from UNK ' s flat , it ' s the night we broke in . & \\
it ' s him , it ' s the man we disturbed . & \\
we can ' t identify him from this . & \\
can ' t see a number plate , even . & \textit{we ' re not going to get rid of
                                      \textbf{him} .} \\
\hline
\hline
 but he was such a puppy ! & \textit{i was so scared .} \\
\hline
how time flies !& \\
it only seems like yesterday since i put 00 candles on his birthday
  cake . & \\
he was such a tall boy !& \\
nearly six feet . & \\
but he was such a puppy ! & \textit{he was a \textbf{young man} !} \\

\hline
\hline
 now just two months ago right here & \textit{and the other ?} \\
\hline
would you swear to that ?& \\
yes , sir , we swear to that . & \\
there you are , an UNK alibi .& \\
serena robbins , you come down here ! & \\
now just two months ago right here & \textit{\textbf{we ' re} on the \textbf{way down} .} \\

\hline
\end{tabular}
\end{center}
\end{table*}

\section{Conclusions}
\label{conclusions}

We studied neural discourse models that can capture long distance
relationships between features found in different utterances of a conversation.
We found that a model with an additional discourse RNN outperforms the
baseline RNN encoder-decoder with an attention mechanism.
Our results indicate that providing more context from previous
utterances improves model performance up to a point.
Qualitative examples illustrate how the discourse RNN produces
increased coherence and cohesion with the rest of the conversation,
while quantitative results based on text mining of discourse markers
show that the amount of deixis, anaphora, and logical consequence
found in the decoder output can be sensitive to the size of the
context window.

In future work, it will be interesting to train discourse models on even
larger corpora and compare conversations in different domains.
By examining the attention weights it should be possible to study what
discourse markers the models are ``paying attention to'' and possibly
provide a powerful new tool for analyzing discourse relationships.
By applying multi-task sequence to sequence learning techniques as in \cite{luong}
we may be able to combine the conversational modeling
task with other tasks such as discourse parsing and/or world knowledge modeling achieve better
overall model performance. Not just for conversations, neural discourse modeling could also be applied to 
written text documents in domains with strong patterns of discourse such as news, legal, healthcare.



\end{document}